\documentclass[letterpaper, 10 pt, conference]{ieeeconf}  

\IEEEoverridecommandlockouts                              

\usepackage{graphicx} 
\usepackage{epsfig} 
\usepackage{amsmath} 
\usepackage{amsfonts}
\usepackage{amssymb}  
\usepackage[caption=false,font=scriptsize,labelfont=scriptsize]{subfig}

\title{\LARGE \bf Minimum Time Strategies for a Differential Drive Robot Escaping  from a Circular Detection Region}

\author{Ubaldo Ruiz
\thanks{U. Ruiz is with the Centro de Investigaci\'on Cient\'ifica y de Educaci\'on Superior de Ensenada (CICESE), 22860, Baja California, M\'exico, {\tt\footnotesize uruiz@cicese.mx}}
}

\begin{document}

\maketitle
\thispagestyle{empty}
\pagestyle{empty}

\begin{abstract}
A Differential Drive Robot (DDR) located inside a circular detection region in the plane wants to escape from it in minimum time. Various robotics applications can be modeled like the previous problem, such as a DDR escaping as soon as possible from a forbidden/dangerous region in the plane or running out from the sensor footprint of an unmanned vehicle flying at a constant altitude. In this paper, we find the motion strategies to accomplish its goal under two scenarios. In one, the detection region moves slower than the DDR and seeks to prevent escape; in another, its position is fixed. We formulate the problem as a zero-sum pursuit-evasion game, and using differential games theory, we compute the players' time-optimal motion strategies. Given the DDR's speed advantage, it can always escape by translating away from the center of the detection region at maximum speed. In this work, we show that the previous strategy could be optimal in some cases; however, other motion strategies emerge based on the player's speed ratio and the players' initial configurations.
\end{abstract}


\section{INTRODUCTION}

An important task in mobile robotics is keeping an intelligent agent under surveillance as it travels in the environment for as much time as possible. Another crucial task is escaping from a stationary region in minimum time. In particular, the first problem can be framed as a pursuit-evasion game with two players having antagonistic goals. In our case, a Differential Drive Robot (DDR) moves on a plane without obstacles, and it is located inside a circular detection region. The DDR's goal is to escape from the region as soon as possible. On the contrary, we assume the detection region can move in any direction, i.e., it is an omnidirectional agent with a speed lower than the one of the DDR, and its objective is to keep the DDR inside for as much time as possible. Note that in the previous setup, the problem of escaping from a stationary region in minimum time can be modeled as an instance where the speed of the detection region is zero. In this work, we formulate the problem as a zero-sum pursuit-evasion game, and using differential games theory, we compute the players' motion strategies to attain their goals. Since the DDR is faster than the detection region, it can always escape by translating away from the center of the detection region. In this work, we found that for some configurations, that strategy corresponds to the time-optimal one; however, other motion strategies emerge based on the players' initial configurations and their speed ratio.

\section{RELATED WORK}

Pursuit-evasion games have been extensively studied in the past \cite{ISAACS-99,MERZ-72,MERZ-74,BERNHARD-77,LEWIN-79,GREENFELD-87}, and still, they receive a lot of attention \cite{STIFFLER-17,CHAUDHARI-21,CARDONA-22,LOZANO-22,LOZANO-22b}; we refer the reader to \cite{CHUNG-11,ROBIN-16} for a taxonomy of pursuit-evasion games. Several works have addressed a pursuit-evasion game involving a DDR and an Omnidirectional Agent (OA) in the literature \cite{MURRIETA-11,RUIZ-13,RUIZ-16,RUIZ-22,RUIZ-23}. In the following paragraph, we describe those works we consider to be closely related to our current work.

The problem of capturing an OA using a DDR in minimum time is addressed in \cite{RUIZ-13}. Like our current work, it is framed as a zero-sum differential game. However, in our case, the DDR wants to leave the OA's detection region, while in \cite{RUIZ-13}, it wants to capture the OA by reaching a given distance. Analogous, in our case, the OA wants to keep the DDR inside its detection region, while in \cite{RUIZ-13}, the OA wants to delay the capture. Those differences imply that the players' motion strategies computed in \cite{RUIZ-13} cannot be used directly in our case. Another study addressing a pursuit-evasion game between an OA and a DDR was presented in \cite{RUIZ-23}. That work investigates the problem of an OA evading surveillance in minimum time from a DDR equipped with a limited field-of-view sensor. The detection region is modeled as a semi-infinite cone. A first distinction between \cite{RUIZ-23} and our current study lies again in the players' goal. In \cite{RUIZ-23}, the DDR plays as pursuer, while in our current setup, it plays as an evader. Analogous, the roles of the OA are also switched. Another difference is the game's terminal condition. In our formulation, the evader escapes by increasing its distance from the pursuer, whereas in \cite{RUIZ-23}, the evader achieves escape by exiting the sides of the cone. Those differences render the motion strategies from \cite{RUIZ-23} inapplicable to our problem. The study of a DDR using a range sensor to surveil an OA is introduced in \cite{RUIZ-22}. In contrast, our study examines a symmetric scenario in which the players' roles are reversed. Note that this change requires computing a new set of motion strategies and their corresponding trajectories since the players' configurations at the end of the game are distinct and the players' goals also differ. Additionally, distinct singular surfaces may appear in the solution. In \cite{BRAUN-2024}, another surveillance game is studied where a non-holonomic evader wants to escape from a moving detection region; however, in their case, the evader is a Dubins Car, which has different time-optimal motion primitives than a DDR, implying that the analysis presented in \cite{BRAUN-2024} is not directly applicable to our problem.

The main contributions of this work are 1) we compute the time-optimal motion strategies for the players and find closed expressions describing their trajectories, 2) we find two singular surfaces in our game: a Transition Surface, where the evader switches controls, and a Dispersal Surface, in which the players have two choices for their controls leading to the same cost, and 3) we present a characterization of the solution based on the players' speed ratio and the detection radius.

\section{PROBLEM FORMULATION}

\begin{figure}[t]
\centering
\subfloat[Realistic space \label{fig:realistic}]{
\includegraphics[width=0.5\linewidth]{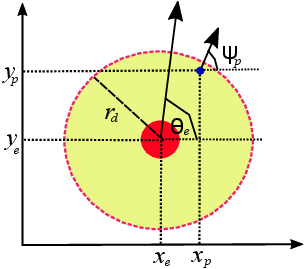}
}\hspace{0.2cm}
\subfloat[Reduced space \label{fig:reduced}]{
\includegraphics[width=0.38\linewidth]{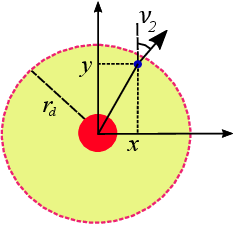}
}
\caption{Game representation. \label{fig:model}} 
\end{figure}

A Differential Drive Robot (DDR), with a maximum translational speed $V_r^{\max}$ is located inside a circular detection region of radius $r_d$ in the plane, and its goal is to escape from it as soon as possible. We assume the circular detection region can move in any direction with speed $V_d^{\max}<V_r^{\max}$, and its objective is to keep the DDR inside of it for as much time as possible. We frame the previous problem as a zero-sum differential game; we denote the DDR as the {\em evader} and the circular detection region as the {\em pursuer}. The game ends when the evader reaches a distance $r_d$ from the pursuer's position, and it can increase it despite the pursuer's resistance. We focus on a pure kinematic problem, where the pursuer's position is represented as $(x_p, y_p) \in \mathbb{R}^2$ and the evader's position by $(x_e, y_e, \theta_e) \in \mathbb{R}^2\times S^1$ (see Fig. \ref{fig:realistic}). Thus, the state of the system is denoted as $\mathbf{x}=(x_p,y_p,x_e,y_e,\theta_e) \in \mathbb{R}^4 \times S^1$, and its evolution as time elapses is described by
\begin{equation}
\label{eq:motioneqs}
\begin{split}
\dot x_p &= v_p \cos \psi_p, \dot y_p = v_p \sin \psi_p, \\
\dot x_e &= \left(\frac{u_1+u_2}{2}\right) \cos \theta_e, \dot y_e = \left(\frac{u_1+u_2}{2}\right) \sin \theta_e, \\
\dot \theta_e &=  \left(\frac{u_2-u_1}{2b}\right),\\
\end{split}
\end{equation}
where $v_p\in [0,V_d^{\max}]$ and $\psi_p\in[0,2\pi)$ are the pursuer's controls. In this case, $u_1, u_2\in[-V_r^{\max},V_r^{\max}]$ are the evader's controls and correspond to the left and right angular velocities of the DDR's wheels. $b$ denotes the distance between the DDR's center and the wheel's location. We call to the previous representation the {\em realistic space}. In this case, all angles are measured in a counter-clockwise direction from the $x$-axis. In this work, we denote as $\rho_v=\frac{V_d^{\max}}{V_r^{\max}}$, to the ratio of the players' speeds, and $\rho_l=\frac{r_d}{b}$ to the ratio between the detection range $r_d$ and the DDR's radius $b$. For the remainder of the paper, we assume $r_d > b$, i.e., the radius of the detection region is bigger than the radius of the DDR evader.

When solving a pursuit-evasion game, using a reference frame fixed to one player is often useful to reduce the state's dimension. In our case, we employ a reference frame fixed to the evader's body, in which the $y$-axis is aligned with its heading direction. We call this representation the {\em reduced space} (see Fig. \ref{fig:reduced}). In this case, all angles are measured in a clockwise direction from the $y$-axis. The following equations describe the coordinate transformation from the realistic space to the reduced space,
\begin{equation}
\label{eq:coordinatetrans}
\begin{split}
x &= (x_p-x_e) \sin \theta_e - (y_p-y_e) \cos \theta_e, \\
y &= (x_p-x_e) \cos \theta_e + (y_p-y_e) \sin \theta_e, \\
v_2 &= \theta_e - \psi_p.
\end{split}    
\end{equation}
The state of the system in the reduced space is denoted by $\mathbf{x}_R=(x,y)\in \mathbb{R}^2$, and its evolution as time elapses is given by
\begin{equation}
\label{eq:motionreduced}
\begin{split}
\dot x &= \left(\frac{u_2-u_1}{2b}\right) y + v_1 \sin v_2, \\
\dot y &= -\left(\frac{u_2-u_1}{2b}\right) x - \left(\frac{u_1+u_2}{2}\right) + v_1 \cos v_2, \\
\end{split}
\end{equation}
where $v_1\in[0,V_d^{\max}]$ and $v_2\in[0,2\pi)$ are the pursuer's controls. Again,  $u_1, u_2\in[-V_r^{\max},V_r^{\max}]$ are the evader's controls. The previous expression is also denoted as $\mathbf{\dot x}_R=f(\mathbf{x}_R,\mathbf{u},\mathbf{v})$, where $\mathbf{u}=(u_1,u_2)$ and $\mathbf{v}=(v_1,v_2)$.

\section{TERMINAL CONDITIONS}

For our problem, the terminal surface $\mathcal{T}$, also known as the target set, corresponds to a circle of radius $r_d$ centered at the evader's body, which in the reduced space is described by the following expression
\begin{equation}
\label{eq:targetset}
\mathcal{T}=\lbrace (x,y) \in \mathbb{R}^2 | x^2+y^2=r_d^2 \rbrace.
\end{equation}
The game ends when $\mathbf{x}_R \in \mathcal{T}$ and the evader can increase the distance between both players regardless of the pursuer's opposition in the next time step. The region of $\mathcal{T}$, where this occurs, is known as the {\em usable part} (UP). That condition is represented by
\begin{equation}
\label{eq:UP}
UP=\lbrace \mathbf{x}_R \in \mathcal{T} | \min_{\mathbf{u}} \max_{\mathbf{v}} \mathbf{n} \cdot f(\mathbf{x}_R,\mathbf{u},\mathbf{v})^T < 0 \rbrace,
\end{equation}
where $\mathbf{n}$ is the inward normal vector to $\mathcal{T}$ at $\mathbf{x}_R$. Note that $\mathcal{T}$ can be parametrized by the angle $s$, thus $\mathbf{n}=(-\sin s, -\cos s)$. Substituting the previous expression and (\ref{eq:motionreduced}) into (\ref{eq:UP}), we have that
\begin{equation}
\label{eq:UP2}
UP= \Bigg\{\min_{u_1, u_2} \Bigg[V_d^{\max} + \left(\frac{u_1+u_2}{2}\right) \cos s < 0 \Bigg] \Bigg\}.
\end{equation}
From (\ref{eq:UP2}), we have that the controls $u_1$ and $u_2$ that minimize that expression must be equal and saturated, and their signs are opposite to the sign of $\cos s$. Recall that $\cos s>0$ for $s\in[0,\frac{\pi}{2})\cup (\frac{3\pi}{2},2\pi)$, thus in this case, $u_1,u_2=-V_r^{\max}$, i.e., the evader translates backward at maximum speed. Analogous, $\cos s < 0$ for $s\in(\frac{\pi}{2},\frac{3\pi}{2})$, hence, $u_1,u_2=V_r^{\max}$, i.e., the evader translates forward at maximum speed. Note that the evader's controls are undefined for $s=\frac{\pi}{2}$ and $s=\frac{3\pi}{2}$; later in the paper, we show that those configurations correspond to a Dispersal Surface. From the previous analysis, the UP contains those configurations in $\mathcal{T}$ where $s\in[0,\arccos \rho_v)\cup (\pi-\arccos\rho_v, \pi+\arccos\rho_v) \cup (2\pi-\arccos\rho_v, 2\pi)$. The {\em boundary of the usable part} (BUP) is given by 
\begin{equation}
\label{eq:BUP}
BUP=\lbrace \mathbf{x}_R \in \mathcal{T} | \min_{\mathbf{u}} \max_{\mathbf{v}} \mathbf{n} \cdot f(\mathbf{x}_R,\mathbf{u},\mathbf{v})^T = 0 \rbrace.
\end{equation}
In our problem, the BUP corresponds to the configurations in $\mathcal{T}$ where $s=\arccos \rho_v$, $s=\pi-\arccos \rho_v$, $s=\pi+\arccos \rho_v$ and $s=2\pi-\arccos \rho_v$.

\section{MOTION STRATEGIES}

In this section, we compute the players' motion strategies to accomplish their goals and the corresponding trajectories in the reduced space. We follow the approach developed by R. Isaacs \cite{ISAACS-99,BASAR-98,LEWIN-12} for solving differential games. The key idea is performing a retro-time integration of the motion equations considering the players' time-optimal controls starting at the configurations in the UP. We denote the retro-time as $\tau=t_f-t$, where $t_f$ is the game's termination time.

To find the players' optimal controls, we construct the Hamiltonian of the system. From \cite{ISAACS-99},
\begin{equation}
\label{eq:hamiltonian}
H(\mathbf{x}_R,\mathbf{\lambda},\mathbf{u},\mathbf{v}) = \mathbf{\lambda}^T \cdot f(\mathbf{x}_R,\mathbf{u},\mathbf{v}) + L(\mathbf{x}_R,\mathbf{u},\mathbf{v}),
\end{equation}
where $\lambda^T$ are the costate variables and $L(\mathbf{x}_R,\mathbf{u},\mathbf{v})$ is the cost function. For problems of minimum time \cite{ISAACS-99}, $L(\mathbf{x}_R,\mathbf{u},\mathbf{v})=1$ . Substituting (\ref{eq:motionreduced}) into (\ref{eq:hamiltonian}), we have that in the reduced space
\begin{equation}
\label{eq:hamiltonian2}
\begin{split}
&H(\mathbf{x}_R, \lambda, \mathbf{u}, \mathbf{v}) = \lambda_{x}\left( \frac{u_2 - u_1}{2b} \right)y  + \lambda_{x} v_1 \sin v_2 \\
&- \lambda_{y}\left( \frac{u_2 - u_1}{2b} \right) x - \lambda_{y}\left( \frac{u_1 + u_2}{2} \right) + \lambda_{y} v_1 \cos v_2 + 1. \\
\end{split}
\end{equation}
For problems of minimum time \cite{ISAACS-99,BASAR-98,LEWIN-12},
\begin{equation}
\label{eq:hamiltonianmax}
\begin{split}
&\min_\mathbf{u} \max_\mathbf{v} H(\mathbf{x}_R, \lambda, \mathbf{u}, \mathbf{v})=0,\\
&u^*=\arg \min_\mathbf{u} H(\mathbf{x}_R, \lambda, \mathbf{u}, \mathbf{v}),\\
&v^*=\arg \max_\mathbf{v} H(\mathbf{x}_R, \lambda, \mathbf{u}, \mathbf{v}),\\
\end{split}
\end{equation}
where $\mathbf{u}^*$ and $\mathbf{v}^*$ denote the player's optimal controls. From (\ref{eq:hamiltonian}) and (\ref{eq:hamiltonianmax}), we have that the evader's controls are
\begin{equation}
\label{eq:evaderctrls}
\begin{split}
&u_1^* = -\mbox{sgn}\left(\frac{-y\lambda_{x}}{b} + \frac{x\lambda_{y}}{b} - \lambda_{y} \right) V_r^{\max}, \\
&u_2^* = -\mbox{sgn}\left( \frac{y\lambda_{x}}{b} - \frac{x\lambda_{y}}{b} - \lambda_{y} \right) V_r^{\max},
\end{split}
\end{equation}
and the pursuer's controls are
\begin{equation}
\label{eq:pursuerctrls}
v_1^* = V_d^{\max},\: \sin v_2^* = \frac{\lambda_{x}}{\gamma},\: \cos v_2^* = \frac{\lambda_{y}}{\gamma},
\end{equation}
where $\gamma = \sqrt{\lambda_{x}^2 + \lambda_{y}^2}$. 

From (\ref{eq:evaderctrls}) and (\ref{eq:pursuerctrls}), one can notice that for computing the players' optimal controls, we require to know the values of $\lambda^T=[\lambda_x \:\: \lambda_y]^T$ as time elapses. To calculate those values, we employ the costate equation. In particular, since we are going to perform a retro-time integration of the motion equations, we utilize the retro-time version of the costate equation
\begin{equation}
\label{eq:costate}
\mathring \lambda =\frac{\partial}{\partial x}H(\mathbf{x}_R, \lambda, \mathbf{u}^*, \mathbf{v}^*),
\end{equation}
where the retro-time derivative of a variable $x$ is represented by $\mathring x$. Substituting (\ref{eq:hamiltonian2}) into (\ref{eq:costate}), and considering the players' optimal controls $\mathbf{u}^*$ and $\mathbf{v}^*$, we have that
\begin{equation}
\label{eq:adjointeq}
\begin{split}
&\mathring \lambda_{x}= -\left( \frac{u_2^*  - u_1^*}{2b}\right) \lambda_{y}, \:\: \mathring \lambda_{y}= \left( \frac{u_2^*  - u_1^*}{2b} \right) \lambda_{x}\\
\end{split}
\end{equation}
We need to find the initial conditions at the game's end ($\tau = 0$) to perform the retro-time integration of (\ref{eq:adjointeq}). At the UP, $x_0=r_d \sin s$ and $y_0=r_d \cos s$. From the traversability conditions, we have that
\begin{equation}
\label{eq:adjointinitial}
\lambda_{x_0} = -\sin s, \:\: \lambda_{y_0} = -\cos s.
\end{equation}
Integrating (\ref{eq:adjointeq}) considering the initial conditions in (\ref{eq:adjointinitial}), and recalling at the game's end, the evader translates at maximum speed; we have that
\begin{equation}
\label{eq:adjointsolutions}
\begin{split}
\lambda_{x} &= -\sin s, \:\: \lambda_{y} = -\cos s.
\end{split}
\end{equation}
Note that $\lambda_x$ and $\lambda_y$ have constant values; thus, the pursuer's motion direction ($v_2^*=s$) in the reduced space is constant. We know that the evader is translating at maximum speed at the end of the game; thus, its motion direction $\theta_e$ in the realistic space is constant. From (\ref{eq:coordinatetrans}), we have that  $\psi_p$, the pursuer's motion direction in the realistic space, is also constant.

In the following paragraphs, we obtain the players' trajectories that lead directly to the terminal conditions; they are known as the {\em primary solution}. To compute them, we require the retro-time version of the motion equations 
\begin{equation}
\label{eq:retroreducedsystem}
\begin{split}
\mathring x &= -\left(\frac{u_2-u_1}{2b}\right) y - v_1 \sin v_2, \\
\mathring y &= \left(\frac{u_2-u_1}{2b}\right) x + \left(\frac{u_1+u_2}{2}\right) - v_1 \cos v_2, \\
\end{split}
\end{equation}
Integrating (\ref{eq:retroreducedsystem}), considering the initial conditions $x_0=r_d \sin s$ and $ y_0=r_d\cos s$ at the UP, and the players' optimal controls, we have that
\begin{equation}
\label{eq:primary}
\begin{split}
x &= \tau V_d^{\max} \sin s  + r_d \sin s,\\
y &= \tau(V_d^{\max} \cos s \mp V_r^{\max}) + r_d \cos s.\\ 
\end{split}
\end{equation}
The sign $-$ is taken if the evader moves backward in the realistic space, and the sign $+$ is taken if it moves forward. Recall that if $s\in[0,\arccos \rho_v)\cup (\pi-\arccos \rho_v,2\pi)$, the evader moves backward at the game's end; otherwise, it is moving forward. (\ref{eq:primary}) provides the system's trajectories in the reduced space. We need to apply the transformation given by (\ref{eq:coordinatetrans}) for computing the corresponding trajectories in the realistic space.

In this game, two behaviors appear when the players follow the trajectories in the primary solution. We found that some trajectories reach the $x$-axis and meet the ones coming from the opposite side, generating a Dispersal Surface on the $x$-axis where the players have two options for their optimal controls, and in others, the evader switches controls before the intersection occurs. The configuration where that change occurs corresponds to a Transition Surface.
 
From (\ref{eq:primary}), we have that the time $t_i$ needed by a primary trajectory to reach the $x$-axis is
\begin{equation}
\label{eq:intersection}    
\tau_i = \left| \frac{r_d \cos s}{V_d \cos s \mp V_r^{\max}} \right|
\end{equation}
where the sign in the denominator is taken as described in the primary solution.
Substituting (\ref{eq:adjointsolutions}) and (\ref{eq:primary}) into (\ref{eq:evaderctrls}) and computing the time $\tau_s$ when the sign functions change their value, we obtain that 
\begin{equation}
\label{eq:switch_time}    
\tau_s=\left|\frac{b\cos s}{V_r^{\max}\sin s}\right|
\end{equation}
\begin{table}[t]
    \caption{Switching control}
    \centering
    \begin{tabular}{c|c}
         Interval & Control  \\
         $s\in(0,\arccos \rho_v)$ & $u_1^*$, from $-V_r^{\max}$ to $V_r^{\max}$. \\
         $s\in(\pi-\arccos \rho_v, \pi)$ & $u_1^*$, from $V_r^{\max}$ to $-V_r^{\max}$. \\
         $s\in(\pi, \pi+\arccos \rho_v)$ & $u_2^*$, from $V_r^{\max}$ to $-V_r^{\max}$. \\
         $s\in(2\pi-\arccos \rho_v, 2\pi)$ & $u_2^*$, from $-V_r^{\max}$ to $V_r^{\max}$. \\
    \end{tabular}
    \label{tab:switches}
\end{table}
Table \ref{tab:switches} summarizes which control switches first for the evader. In our game, we have that the primary trajectories reach the $x$-axis if $\tau_i<\tau_s$; otherwise, the evader switches its controls and a new integration of (\ref{eq:adjointeq}) and (\ref{eq:retroreducedsystem}) is required to compute the trajectories emanating from the Transition Surface. We denote as $\lambda_{x_s}$, $\lambda_{y_s}$, $x_s$ and $y_s$ to the values of $\lambda_x$, $\lambda_y$, $x$ and $y$ at $\tau_s$. The retro-time integration of (\ref{eq:adjointeq}) with those initial conditions give us
\begin{equation}
\label{eq:adjointsolutions2}
\begin{split}
\lambda_{x} = -\sin\left[s - \left(\frac{u_2^*-u_1^*}{2b}\right)(\tau-\tau_s)\right] \\
\lambda_{y} = -\cos\left[s - \left(\frac{u_2^*-u_1^*}{2b}\right)(\tau-\tau_s)\right]
\end{split}
\end{equation}
for $\tau \geq \tau_s$. The retro-time integration of (\ref{eq:retroreducedsystem}) with the previous initial conditions leads to
\begin{equation}
\label{eq:rotation}
\begin{split}
x(\tau) = &-y_s\sin\left[\left(\frac{u_2^*-u_1^*}{2b}\right)(\tau-\tau_s)\right] \\
&+x_s\cos\left[\left(\frac{u_2^*-u_1^*}{2b}\right)(\tau-\tau_s)\right]\\
&-(\tau-\tau_s) V_d^{\max}\sin\left[s-\left(\frac{u_2^*-u_1^*}{2b}\right)(\tau-\tau_s)\right]\\
y(\tau) = &x_s\sin\left[\left(\frac{u_2^*-u_1^*}{2b}\right)(\tau-\tau_s)\right] \\
&+y_s\cos\left[\left(\frac{u_2^*-u_1^*}{2b}\right)(\tau-\tau_s)\right]\\
&-(\tau-\tau_s) V_d^{\max}\cos\left[s-\left(\frac{u_2^*-u_1^*}{2b}\right)(\tau-\tau_s)\right]
\end{split}
\end{equation}
We have that for $\tau\geq \tau_s$ the evader is rotating in place, and its motion direction in the realistic space is given by
\begin{equation}
\label{eq:thetarot}
\theta_e = \theta^s_e - \left( \frac{u_2^*-u_1^*}{2b} \right)(\tau - \tau_s)
\end{equation}
where $\theta^s_e$ is the evader's initial motion direction in the realistic space. 
Substituting (\ref{eq:adjointsolutions2}) into (\ref{eq:pursuerctrls}), we have that
\begin{equation}
\label{eq:v2rot}
v_2^* = s - \left(\frac{u_2^*-u_1^*}{2b}\right)(\tau - \tau_s)
\end{equation}
Now, substituting (\ref{eq:thetarot}) and (\ref{eq:v2rot}) into (\ref{eq:coordinatetrans}) we obtain $\psi_p = \theta^s_e - s$, the pursuer's motion direction in the realistic space. Note that $\psi_p$ is a constant value; thus, the pursuer is following a straight line in the realistic space.

We found that the previous trajectories reach the $x$-axis and meet the ones coming from the opposite side, generating a Dispersal Surface on the $x$-axis. On the Dispersal Surface, the players have two options for their optimal controls; the evader can rotate in place either clockwise or counterclockwise, while the pursuer can go to the upper or lower part of the UP. However, each player must choose its controls correctly, otherwise, it will benefit the other one. Thus, they face the dilemma of knowing the control applied by the other to make its choice.

An important question addressed in pursuit-evasion games is solving the decision game, i.e., determining under which conditions each player wins the game. Note that in this case, since the evader is faster than the pursuer, it can always escape. However, even in such cases, it is important to compute the barrier trajectory \cite{ISAACS-99,BASAR-98,LEWIN-12}, since it may imply that the players need to perform a particular motion strategy \cite{MERZ-74, RUIZ-12}. A retro-time integration of the motion and adjoint equations starting at the BUP is done to find such a trajectory. In our game, we encountered that it immediately exited the playing space. Thus, it does not require computing an additional strategy.

\section{CHARACTERIZATION OF THE SOLUTION}

\begin{figure}[t]
\centering
\subfloat[$\rho_v=0.2$ \label{fig:p1e.2l4}]{
\includegraphics[width=0.45\linewidth]{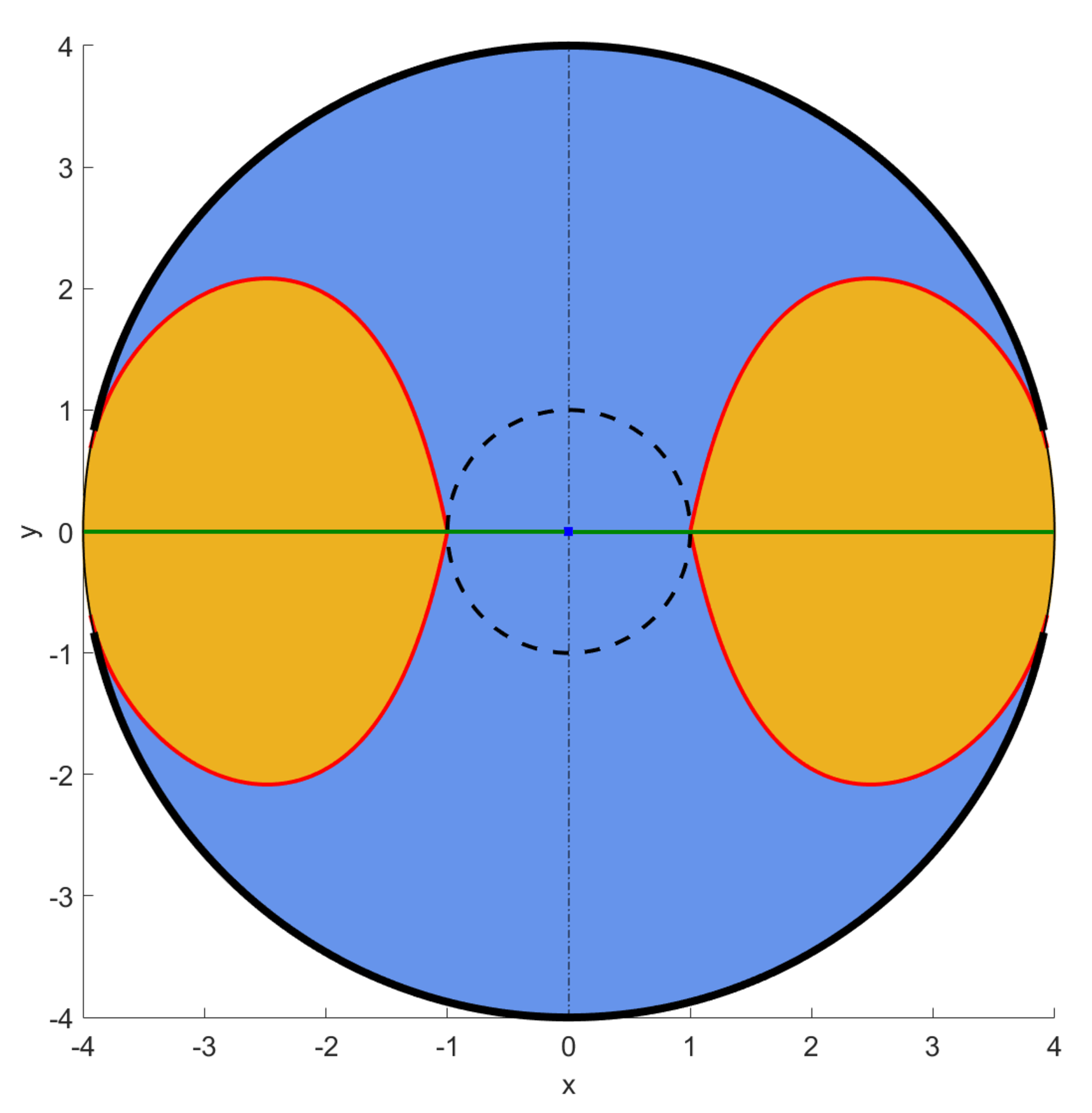}
}
\subfloat[$\rho_v=0.4$ \label{fig:p1e.4l4}]{
\includegraphics[width=0.45\linewidth]{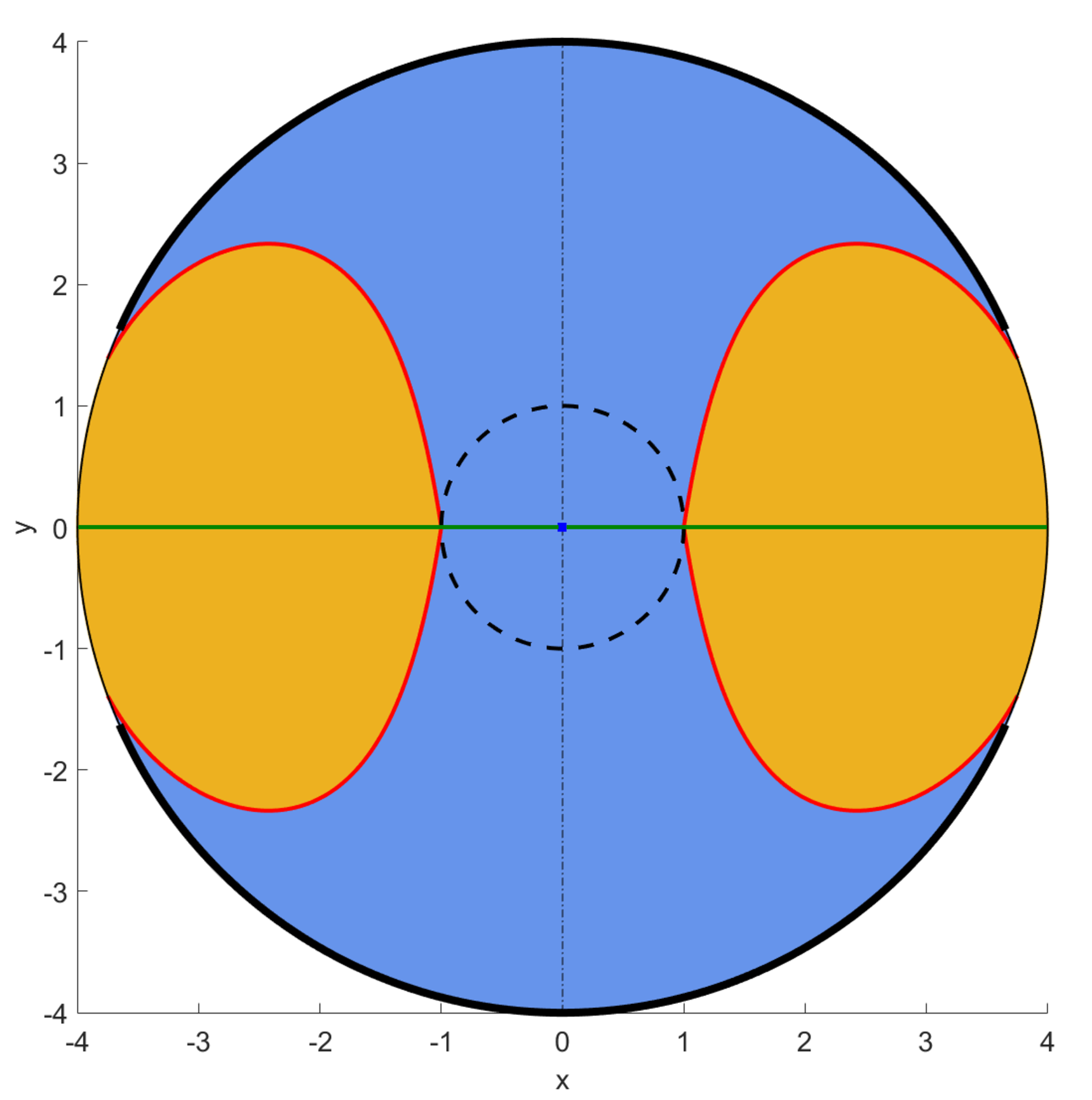}
}\\
\subfloat[$\rho_v=0.6$ \label{fig:p1e.6l4}]{
\includegraphics[width=0.45\linewidth]{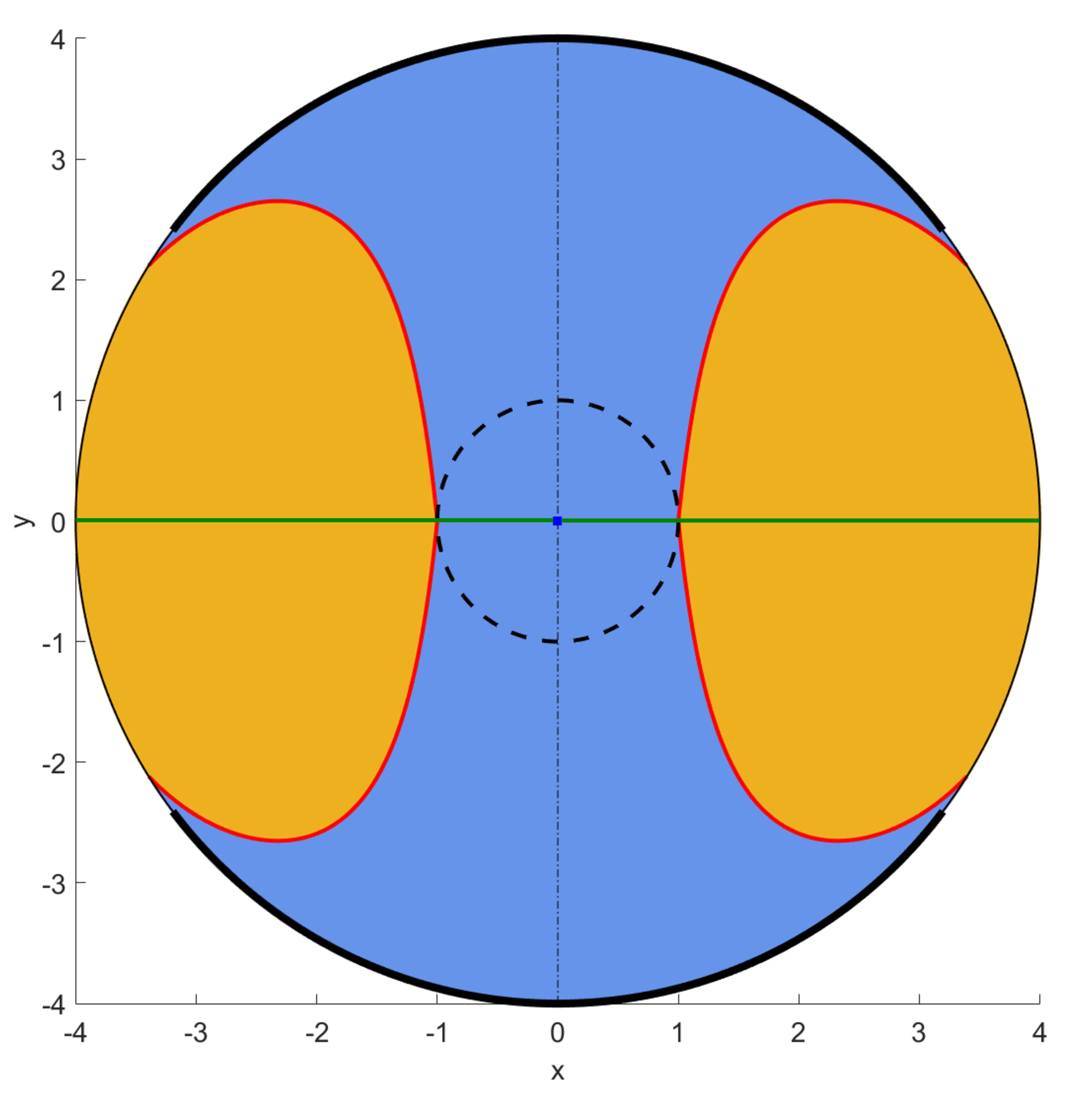}
}
\subfloat[$\rho_v=0.8$ \label{fig:p1e.8l4}]{
\includegraphics[width=0.45\linewidth]{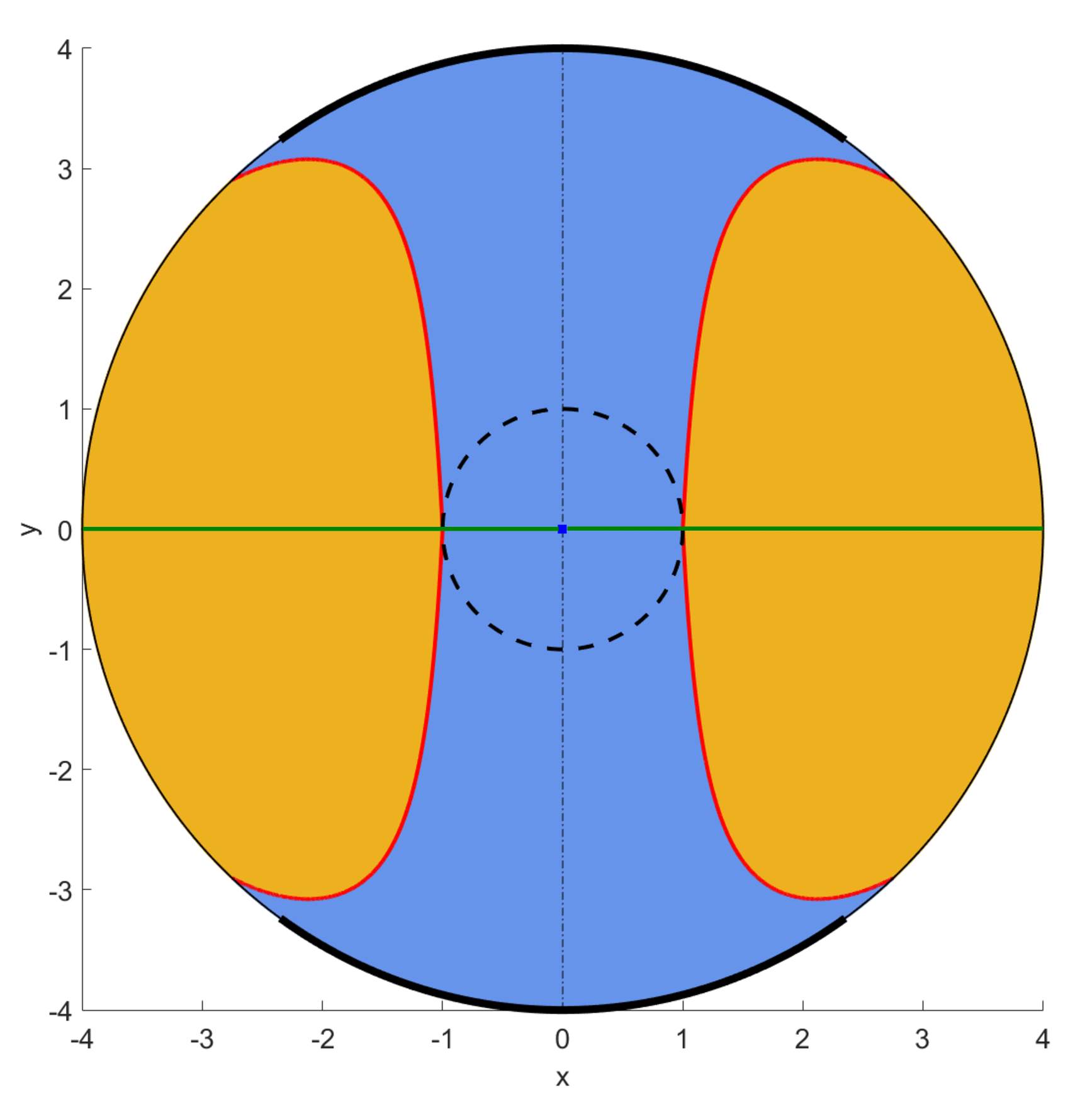}
}
\caption{Partition of the reduced space for different values of $\rho_v$. The black circle represents the detection region, the bold arcs indicate the UP, and the dashed circle is the evader's radius. The blue region corresponds to the configurations where the system follows a trajectory of the primary solution. Similarly, the golden region indicates those configurations where the evader performs a rotation in place. The red curve corresponds to the Transition Surface, those configurations where the evader switches its control. The green line indicates the Dispersal Surfaces at the $x$-axis. \label{fig:vrhov}} 
\end{figure}

\begin{figure}[h]
\centering
\subfloat[$\rho_l=2$ \label{fig:p1e0l2}]{
\includegraphics[width=0.45\linewidth]{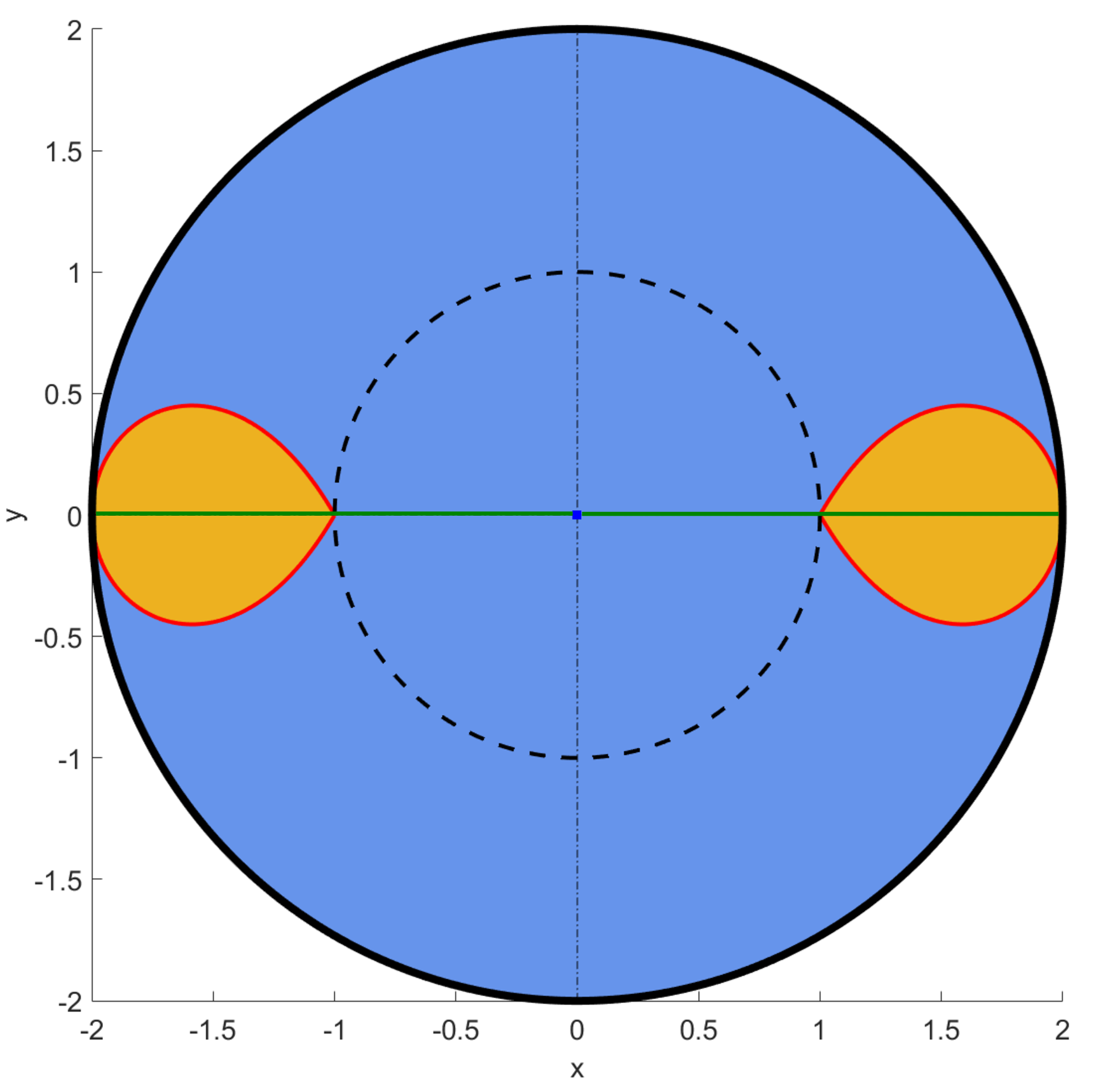}
}
\subfloat[$\rho_l=4$ \label{fig:p1e0l4}]{
\includegraphics[width=0.45\linewidth]{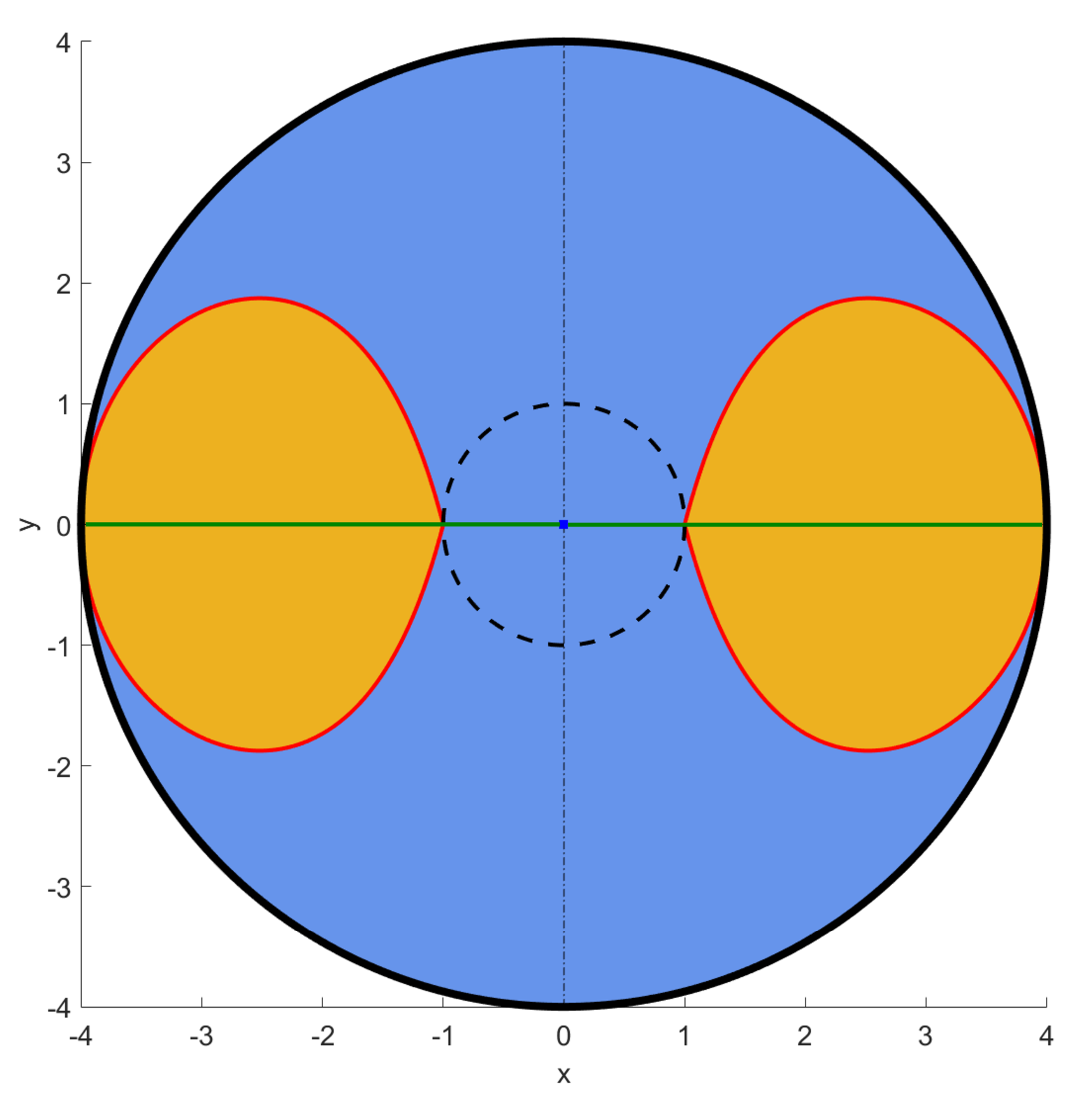}
}\\
\subfloat[$\rho_l=6$ \label{fig:p1e0l6}]{
\includegraphics[width=0.45\linewidth]{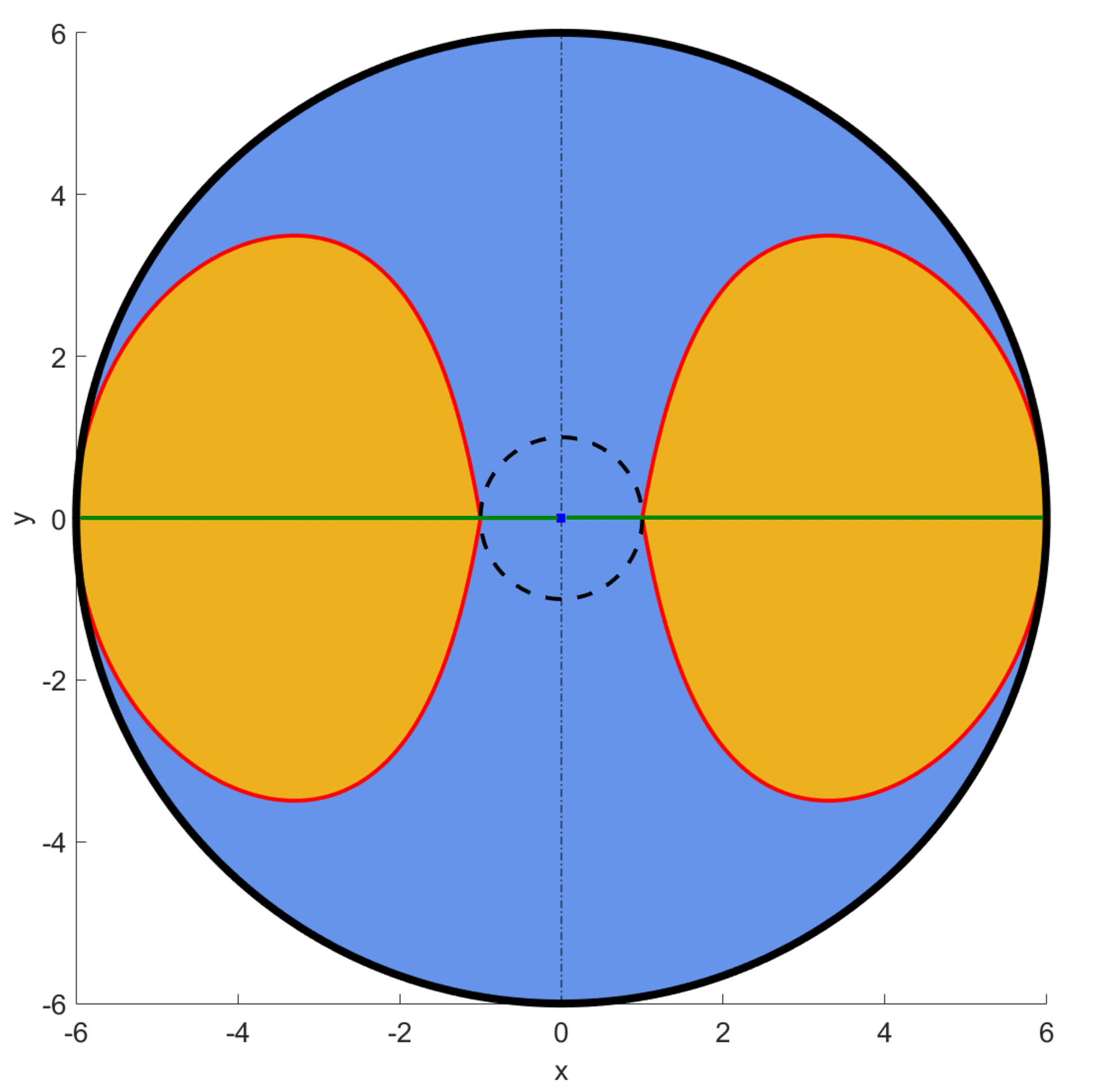}
}
\subfloat[$\rho_l=8$ \label{fig:p1e0l8}]{
\includegraphics[width=0.45\linewidth]{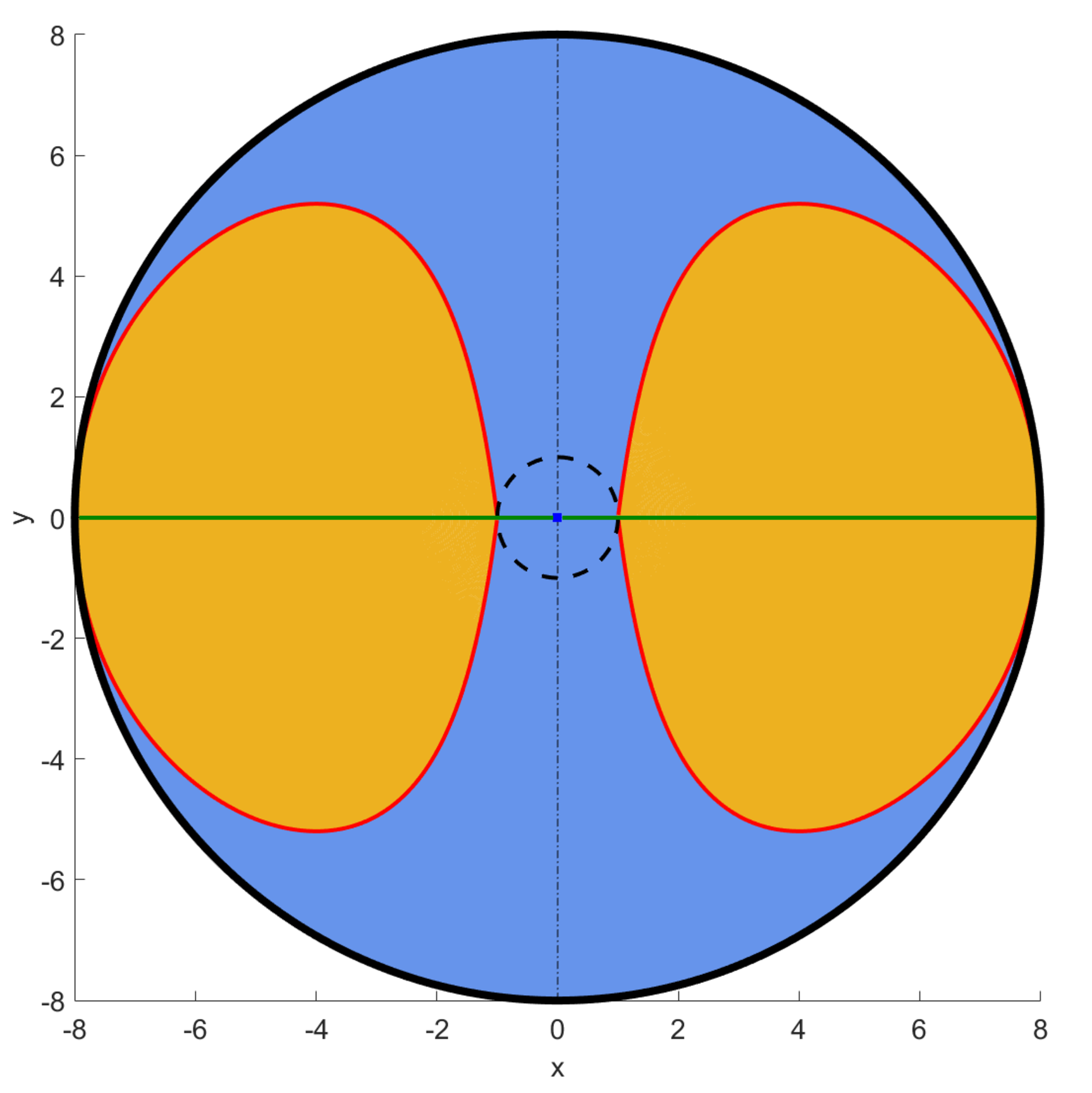}
}
\caption{Partition of the reduced space for different values of $\rho_l$. 
\label{fig:vrhol}} 
\end{figure}

In this section, we present a characterization of the solution for different values of $\rho_v$ and $\rho_l$, see Figs. \ref{fig:vrhov} and \ref{fig:vrhol}. As was described in the previous section, we have two types of motion strategies in the reduced space. In the first one, known as the primary solution (shown in blue) and emanating directly from the UP, both players translate at maximum speed. In the second one, the evader performs a rotation in place while the pursuer translates at maximum speed (shown in golden color). In this work, we found that depending on the values $\rho_v$ and $\rho_l$, the regions of the playing space associated with the previous motion strategies change their size. 

Figs. \ref{fig:vrhov} and \ref{fig:vrhol}, illustrate those changes considering two setups. In the first one, $\rho_l=4$ while $\rho_v=0.2, 0.4, 0.6,0.8$. In Fig. \ref{fig:vrhov}, we can observe that as the pursuer becomes faster, the region (golden color) where the evader performs a rotation in place becomes larger. This may appear a counter-intuitive result since the evader cannot increase the distance from the pursuer while it rotates at maximum speed. However, note that as $\rho_v$ increases, the bold arc displaying the UP reduces its length, indicating that the pursuer's motion is more aligned with the evader's heading when the escape occurs. Note that as the pursuer becomes faster, increasing the distance from it also becomes more difficult for the evader.

In the second setup, $\rho_l=2,4,6,8$ while $\rho_v=0$. Note that this corresponds to the case where the detection region is stationary and also allows the study of the influence of its radius in the game. In Fig. \ref{fig:vrhol}, we can observe that as $\rho_l$ increases, the region (golden color) where the evader performs a rotation in place also increases its size. Note that the bold arc indicating the UP almost covers the whole boundary of the detection region. 

\section{SIMULATIONS}

\begin{figure}[t]
\centering
\includegraphics[scale=0.2]{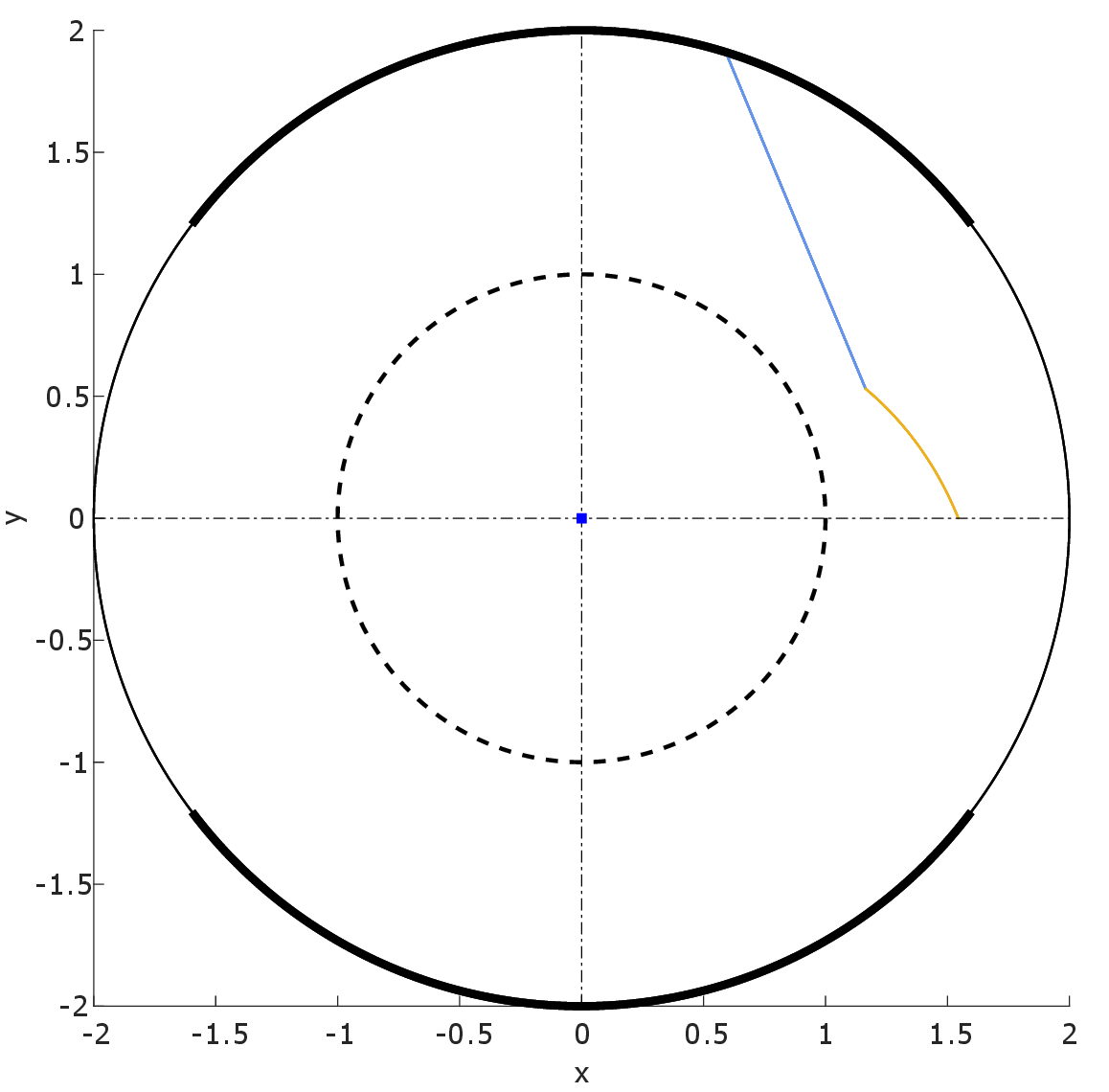}
\caption{Trajectory of the system in the reduced space. The black circle represents the detection region, the bold arcs indicate the UP, and the dashed circle is the evader's radius.  The blue line corresponds to the portion belonging to the primary solution, and the golden curve to the portion where the evader performs a rotation in place.
}
\label{fig:reducedsim1}
\end{figure}

\begin{figure}[t]
\centering
\includegraphics[scale=0.32]{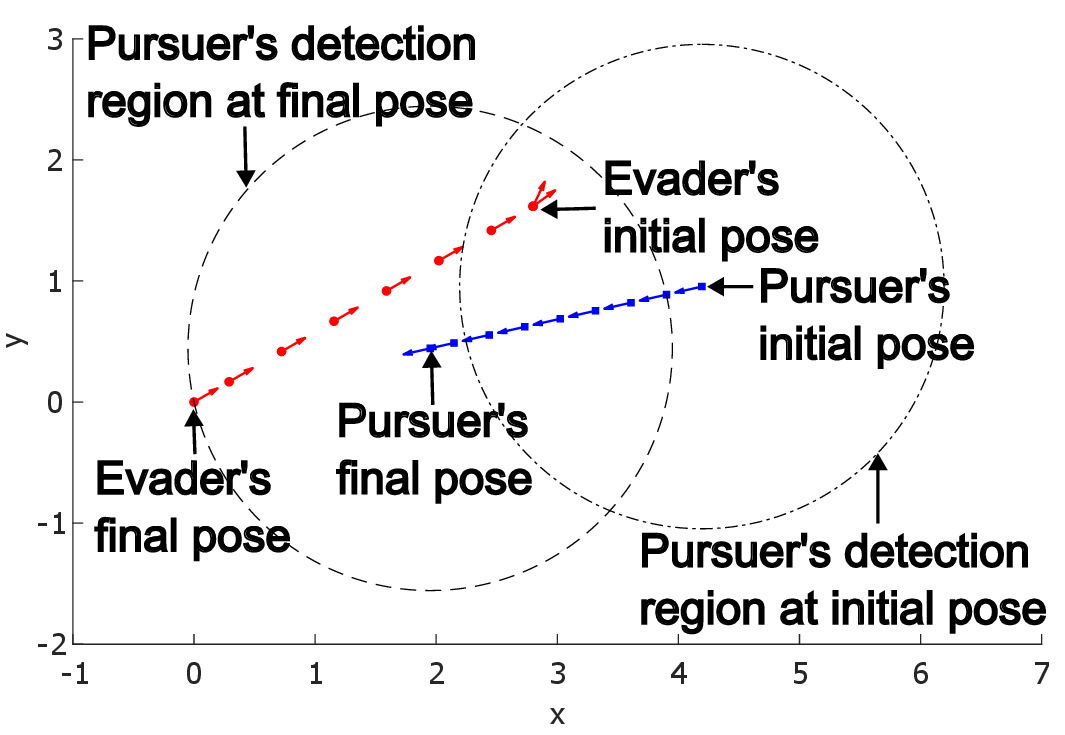}
\caption{Trajectory of the players in the realistic space. The blue squares indicate the pursuer's positions, and the blue arrows its motion directions. Similarly, the red dots correspond to the evader's positions, and the red arrows its motion directions. The circles indicate the detection region at the initial and final players' configurations.}
\label{fig:realisticsim1}
\end{figure}

\begin{figure}[t]
\centering
\subfloat[$\tau=0$ \label{fig:snapshot1}]{
\includegraphics[width=0.45\linewidth]{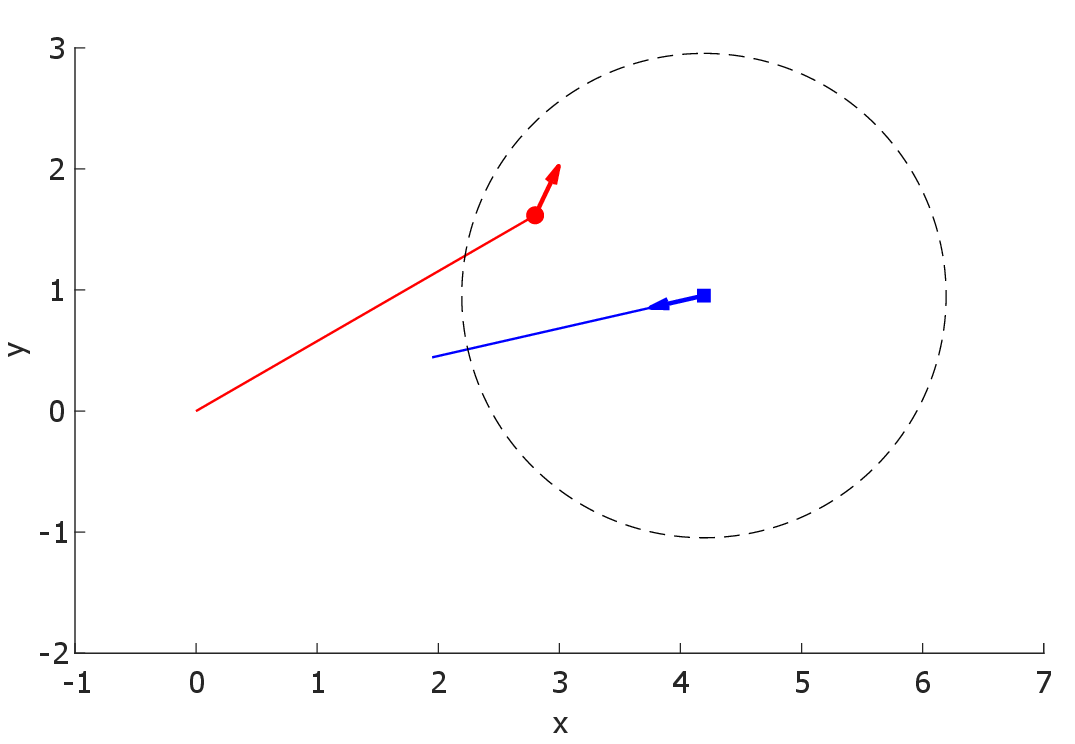}
}
\subfloat[$\tau=1s$ \label{fig:snapshot2}]{
\includegraphics[width=0.45\linewidth]{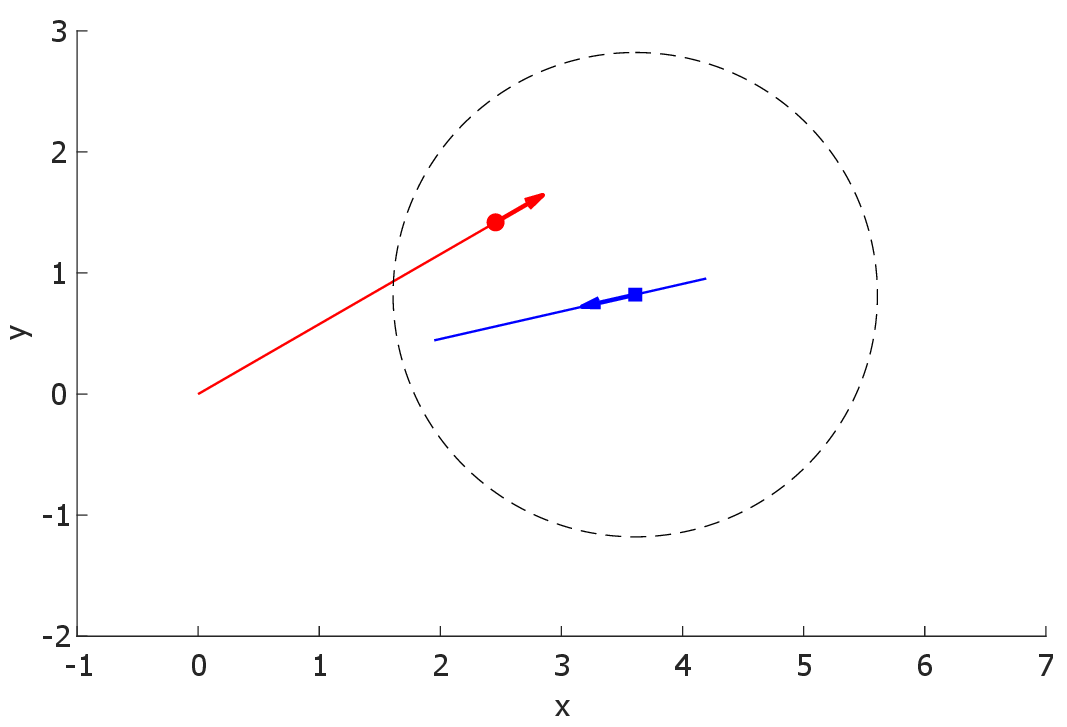}
}\\
\subfloat[$\tau=2s$ \label{fig:snapshot3}]{
\includegraphics[width=0.45\linewidth]{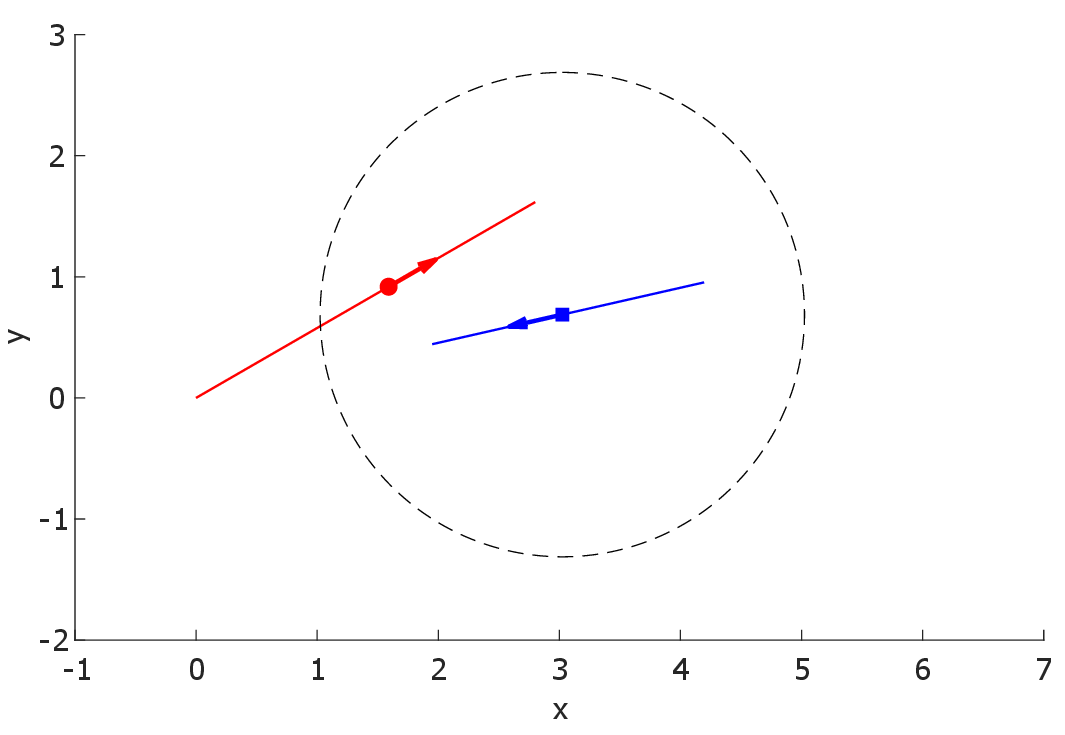}
}
\subfloat[$\tau=3.84s$ \label{fig:snapshot5}]{
\includegraphics[width=0.45\linewidth]{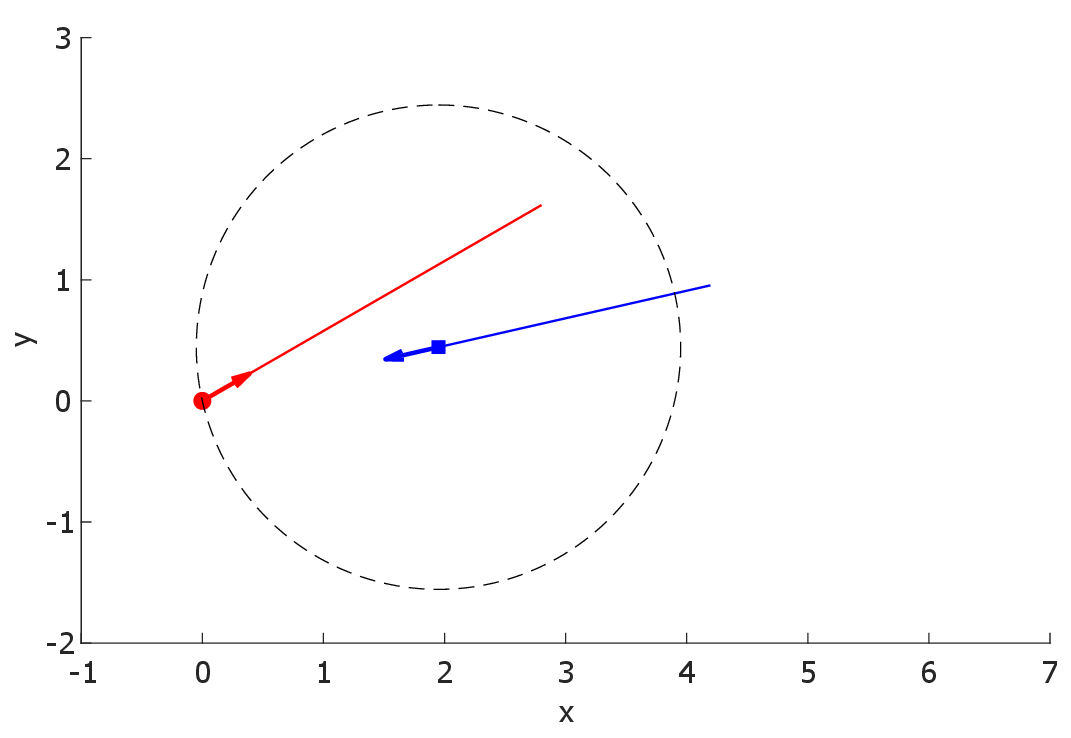}
}
\caption{Snapshots of the players' configurations in the realistic space. The pursuer is represented by the blue square and the evader by the red dot. The dashed circle indicates the detection region centered at the evader's position. The blue curve indicates the entire pursuer's trajectory, and the red curve is the evader's one. The elapsed time is indicated for each figure.\label{fig:snapshots}} 
\end{figure}

In this section, we present a numerical simulation of the players' motion strategies in the reduced and realistic spaces. The parameters of the simulation are $V_r^{\max}=1m/s$, $V_d^{\max}=0.6m/s$, $b=1m$, $s=0.3rad$  and $r_d=2m$. Fig. \ref{fig:reducedsim1} shows the trajectory of the system in the reduced space. The blue line corresponds to the portion belonging to the primary solution, and the golden curve to the portion where the evader performs a rotation in place. In the realistic space, see Fig. \ref{fig:realisticsim1}, we can observe that the evader is initially located inside the detection region and starts its motion rotating in place at maximum speed. At the same time, the pursuer translates at maximum speed. After some time, the evader switches its controls and performs a backward translation at maximum speed. The pursuer continues following a straight-line trajectory until the evader reaches the boundary of the detection region and escapes. Fig. \ref{fig:snapshots} shows several snapshots of the players' configurations in the realistic space as they perform their goals.

\section{CONCLUSIONS}

In this work, we addressed the surveillance problem of keeping a Differential Drive Robot inside a circular detection region for as much time as possible as it moves on the environment. We consider two setups, one in which the detection region can move in any direction and another in which is fixed. We framed the problem as a zero-sum pursuit-evasion game with two players having antagonistic goals, and we computed the time-optimal motion strategies for the players in closed form. In particular, we exhibited the existence of two singular surfaces: a Transition Surface, where the evader switches controls, and a Dispersal Surface, in which the players have two choices for their controls, leading to the same cost. We presented a characterization of the solution based on the players' speed ratio and the detection radius. As future work, we are interested in developing motion strategies for two or more pursuers that cooperate to maintain surveillance.

\addtolength{\textheight}{-12cm}   


\bibliographystyle{IEEEtran}
\bibliography{manuscript}

\end{document}